\documentclass{article}

    \PassOptionsToPackage{numbers, compress}{natbib}

     \usepackage[final]{neurips_2023}

\usepackage{algorithm}
\usepackage{algpseudocode}
\usepackage[utf8]{inputenc} 
\usepackage[T1]{fontenc}    
\usepackage{hyperref}       
\usepackage{url}            
\usepackage{booktabs}       
\usepackage{amsfonts}       
\usepackage{amsmath}       
\usepackage{multirow}
\usepackage{nicefrac}       
\usepackage{microtype}      
\usepackage{xcolor}         
\usepackage{graphicx}
\usepackage{subcaption}

    \title{Cognitively Inspired Cross-Modal Data Generation Using Diffusion Models}

\author{%
  Zizhao Hu\\
  University of Southern California \\
  \texttt{zizhaoh@usc.edu} \\  
   \And
   Mohammad Rostami \\
   University of Southern California \\
   \texttt{rostamim@usc.edu} \\
}

\begin{document}

\maketitle

\begin{abstract}
 Most existing cross-modal generative methods based on diffusion models use guidance to provide control over the latent space to enable conditional generation across different modalities. Such methods focus on providing guidance through separately-trained models, each for one modality. As a result, these methods suffer from cross-modal information loss and are limited to unidirectional conditional generation. Inspired by how humans synchronously acquire multi-modal information and learn the correlation between modalities, we explore a multi-modal diffusion model training and sampling scheme that uses channel-wise image conditioning to learn cross-modality correlation during the training phase to better mimic the learning process in the brain. Our empirical results demonstrate that our approach can achieve data generation conditioned on all correlated modalities. 
\end{abstract}

\section{Introduction}

Natural intelligent agents perceive the  external world via processing multimodal sensory inputs, e.g., vision and sounds. Unified multimodal processing allows natural agents to gain a more robust and reliable perception of the world. The data modalities are often correlated with each other and together play a complementary role because they describe the same world. As a result, conditional cross-modal inference serves a crucial function in intelligent agents to complement deficiencies in sensory inputs. For example, Pavlov's experiment~\cite{pavl}   where a dog salivates at a ringing sound suggests that a stimulus can be triggered by a previously experienced correlated stimulus. This phenomenon is connected to the associative learning happening in the hippocampus~\cite{hipo, hihe, hila}, where the association is learned using the Hebbian process~\cite{hebb,triche2022exploration} and experience replay~\cite{pfeiffer2020content,rostami2020generative}. This process makes the brain prone to reactivate   full activation patterns, given a partial activation, allowing multimodal conditional memory generation~\cite{hime}.
In contrast,   generative AI models are primarily developed to process unimodal data, and generative models for  cross-modal data generation are limited and naive. 

VAEs~\cite{vae} explicitly model the data distribution of a single modality.  VAEs can be generalized to a multimodal scenario, where  the joint distribution of multiple modalities is learned in the latent space.  Conditional cross-modal   data generation is then possible using the Bayes' rule~\cite{khattar2019mvae,rostami2023overcome,hu2023encoding}. However, this  modeling approach  deviates from how the brain learns to generate cross-modal data. Moreover, modeling  distributions of complex domains explicitly is a challenging task. GAN~\cite{gan} and guided score-based diffusion models~\cite{Diffusion, DDPM, SMLD} bypass the explicit modeling of the data distribution by modeling conditional probabilities. Both approaches often create more realistic samples compared to VAEs. In addition to being more efficient in text/image generation tasks, implicit modeling  allows the incorporation of conditional probability to guide the data generation process. Note, however, a separate  classifier and a large labeled dataset are needed to achieve conditional data generation. 

\begin{figure}[t!]
\centering
\includegraphics[width=\linewidth]{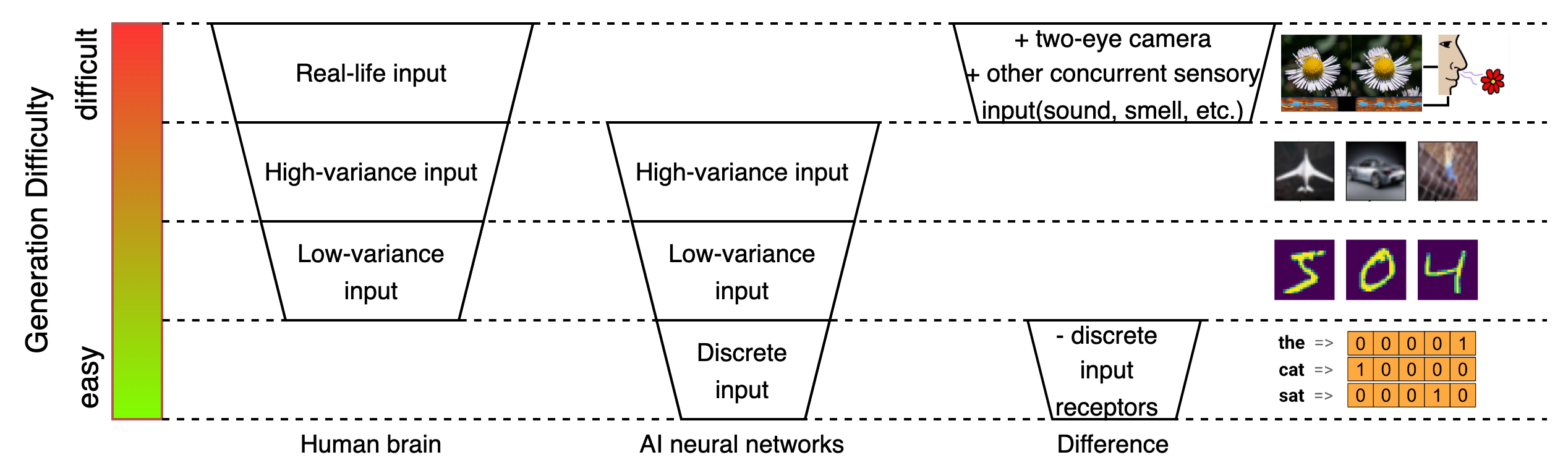}
\caption{Generation difficulty in the human brain versus AI: Input types are different between the human brain and the current AI systems. Each of the input types has a different level of generation difficulty. Existing generative AI systems trade-off real-life multimodal input for generation simplicity, while humans trade off accurate generation for better learning of cross-modality correlations in real-life multimodal data. This leaves a gap between human intelligence and AI, which we aim to close.)}
\label{fig:diff}
\end{figure}

In addition, when it comes to language-conditioned image generation and autoregressive language generation in AI, predefined discrete embeddings are used as the language encoding. However, humans acquire language through associative learning over an extended training period. A child is exposed to all other modalities that happen concurrently with words' pronunciations. Later in life, the connection between visual texts and word pronunciation is further enhanced. All the language inputs come in noisy forms, and the denoised concepts are formed through multimodal learning. This learning process includes multiple modalities which have a large discrepancy with current separately pre-trained unimodal AI schemes in terms of difficulty and input types (see Figure    \ref{fig:diff}).

Due to the above  discrepancies between the training schemes of AI and natural intelligence, current AI systems cannot achieve the same performance similar to the brain in cross-modal data generation. To close this gap, we need to develop a generative AI scheme such that: 
(i)   cross-modal knowledge is acquired during training via a single model, (ii)   all correlated modalities are generated in all directions in the embedding space, (iii) be robust against noisy language inputs, and (iv)   a single continuously trained neural network is used to achieve these capabilities. The lack of these capabilities in the existing generative AI approaches  motivates us to explore the possibility of developing a more human-like multimodal generative AI scheme equipped with these capabilities. More specifically, we focus on  score-based diffusion models~\cite{Diffusion, DDPM, SMLD} and adopt them for our purpose. These models have demonstrated superior performances in image generation tasks. When denoising sampling is guided in these models, conditional data generation also becomes feasible~\cite{Classifier, Classifierfree}. By further guiding these models with pre-trained text encoders, the correlation between text and image can be captured, leading to several   text-to-image generation applications~\cite{DALLE, Imagen}. These features and results support the idea that diffusion models can be   adapted to model human learning better.

We close the gap between human-like multimodal learning and   AI   by exploring a channel-wise image-guided\footnote{Different modalities are transformed and represented as different channels in the image data, analogous to how color images have separate red, green, and blue channels.} multimodal joint diffusion training scheme that transforms and aligns multimodal data in the input space rather than in the learned latent space. By doing so, we develop  more adaptable and robust models that better mimic human learning processes.
Our contributions include:
\begin{enumerate}
    \item By transforming all input modalities   into channel-concatenated spatial inputs, we enable \textit{multi-directional generation}\footnote{Conditional generation between all possible combinations of all correlated modalities, e.g., text-to-image, image-to-audio, text-and-audio-to-image, etc.} across all   modalities using a single generative neural network.
    \item Our model trains on noisy data in all modalities, allowing for \textit{noisy language inputs} as conditions in the process of conditional generation,   enhancing  robustness in diverse scenarios.
    \item Our model provides a learning scheme that is \textit{empirically similar to biological neural networks}, aiming to inspire further research in this intriguing intersection of fields.

\end{enumerate}

\section{Preliminaries and Problem definition}

We briefly explain the score-based diffusion learning process and then define our problem.

\subsection{Score-Based Diffusion Models}
Score-based diffusion models~\cite{Diffusion, DDPM, SMLD}  are a class of generative models that learn to generate new samples from a target data distribution by approximating the score function. The score function is defined as the gradient of the log-probability density of the training data points:
\begin{equation}
S(x) = \nabla_x \log p(x).
\end{equation}

The score function of a timestamped diffusion process is approximated  using a suitable neural network, denoted as $s_\theta(\cdot)$:
\begin{equation}
S(x) \approx s_\theta(x_t),
\end{equation}

The training  variational lower-bound (VLB) objective is approximated to be the MSE  discrepancy between the true score function and the approximation given by the neural network:
\begin{equation}
L(\theta) = \mathbb{E}[(S(x) - s_\theta(x_t))^2].
\end{equation}

By minimizing the MSE, the neural network learns the score function.
Modeling the score function allows the model to generate the data distribution by taking small steps along the gradient direction.

\subsubsection{Score-Based Diffusion Model Training Process}
Given a dataset $\mathcal{D}$ of independent and identically distributed (i.i.d.) samples from a data distribution $p_{data}(x)$, the training process of the score-based diffusion model involves two steps:
\begin{enumerate}

\item  \textbf{Noise addition process}: Given a data sample $x$, we add progressively more Gaussian noise over $T$ steps to generate a noisy observation $x_T$. The process can be formalized as follows:

\begin{equation}
x_t = \sqrt{1 - \beta_t} x_{t-1} + \sqrt{\beta_t} \xi_t,
\end{equation}

where $\xi_t \sim \mathcal{N}(0, I)$, $t=1,ldots,T$, $x_0 = x$, and $\{\beta_t\}_{t=0}^{T}$ is a pre-specified noise schedule.

\item  \textbf{Score function learning}: The   function $s_\theta(x_t, t)$ approximates the gradient of the log-density of the distribution $p(x_t | x)$. The parameters $\theta$ of the score function are learned as follows:

\begin{equation}
\mathcal{L}(\theta) = \mathbb{E}_{p_{data}(x)}\left[ \frac{1}{2T} \sum_{t=0}^{T-1} \left\| s_\theta(x_t, t) + \nabla_{x_t} \log p(x_t | x) \right\|^2 \right].
\end{equation}
\end{enumerate}
\subsubsection{Score-based Diffusion Model Sampling  Process}

The sampling (inference) process of the score-based diffusion model is a denoising process in reverse:

\begin{enumerate}
    
\item \textbf{Initial noisy observation}: We first sample an initial noisy observation $x_T$ from a simple distribution, such as a standard Gaussian distribution to start from.

\item \textbf{Denoising process}: We then apply the learned score function to denoise $x_T$ over $T$:

\begin{equation}
x_{t-1} = \frac{1}{\sqrt{1 - \beta_t}} (x_t - \frac{\beta_t} {\sqrt{1-\bar{\beta_t}}}s_\theta(x_t, t)) + \sqrt{\beta_t}\xi,
\end{equation}

where $t$ ranges from $T$ to $0$, and the final sample $x_0$ is outputted as a data sample.

\end{enumerate}

\subsection{Guided diffusion models}
We can condition the score function on a   label $y$ to learn conditional generative models. This objective can be done by leveraging the Bayes' rule to decompose the conditional score function~\cite{dhariwal2021diffusion}:
\begin{equation}
S(x|y) = \nabla_x \log p(x|y) = \nabla_x \log p(y|x) + \nabla_x \log p(x) 
\label{gfm}
\end{equation}

When $y$ is independent of $x$ during the generation process, the third term is zero.

\textbf{Classifier guidance}: By providing a weight term $\omega$ to adjust the balance between the unconditional score function and the classifier guidance during the sampling phase, we get:
\begin{equation}
S(x|y) = \omega \underbrace{\nabla_x \log p(y|x)}_{\text{classifier guidance}} + \underbrace{\nabla_x \log p(x)}_{\text{unconditional score function}}
\end{equation}

\textbf{Classifier-free guidance}: We can also model the unconditional score function $\nabla_x \log p(x)$ and the joint score function $\nabla_x \log p(x, y)$ at the same time, to substitute the classifier guidance~\cite{ho2021classifier}:
\begin{equation}
S(x|y) = \omega (\nabla_x \log p(x,y) - \nabla_x \log p(x)) + \nabla_x \log p(x)
\end{equation}

 \subsection{Problem Definition}

Given multimodal data $X$, our goal is to train a model such that when we are presented with a subset $x' \in X$ of partially observed modalities, we can generate the unobserved modalities $x = X - x'$. Humans are able to do this task relatively well. For example, when a portion of a scene occluded but we hear a relevant sound, we can speculate what is the visual cue behind the occlusion. Here $x$ and $x'$ can be any combination of complementary subsets. The simplest form can be text-to-image, and image-to-text, when we only consider two modalities. Current cross-modal generation approaches are limited to uni-directional generation, e.g., $x'$ is restricted to text. 

Additionally, the observed subset $x'$ is subject to noise in real-life scenarios. As a result, the generation of $x$ should be relatively robust against the input noise. Current text-to-image models require exact text embeddings, and cannot perfectly mimic this natural constraint.

\section{Proposed Method}

Inspired by classifier-free guidance and how humans cognitively generate cross-modality data given partial modalities' inputs, we   modify the conditional score function in equation \eqref{gfm}:
\begin{equation}
S(X|x') = \nabla_X \log p(X|x') = \nabla_X \log p(x'|X) + \nabla_X \log p(X),
\label{cfg}
\end{equation}
where $x'$ is a component of all concurrent modalities in the input $X$. 

We  can use the joint probability distribution $p(X) = p(x, x')$ and simplify Eq.~\eqref{cfg} as:
\begin{equation}
S(x, x'|x') = \nabla_{(x, x')} \log p(x, x'|x') = \nabla_{(x, x')} \log p(x'|x, x') + \nabla_{(x, x')} \log p(x, x'),
\end{equation}
which then reduces to the following:
\begin{equation}
S(x|x') = \nabla_{(x, x')} \log p(x|x') = \nabla_{(x, x')} \log p(x, x').
\end{equation}
This modeling of a conditional probability distribution using joint probability distribution is possible due to the score-matching learning scheme. To model the joint probability score $\nabla_{(x, x')} \log p(x, x')$, we choose the joint diffusion process of the channel-wise concatenation of $x$ and $x'$, where ``channel'' is the third dimension usually used as RGB channels in colored image input. Since $x$ and $x'$ span different modalities, a lossless conversion to a single channel-compatible modality is required. We choose to use a multi-channel 2-dimensional array similar to RGB images, where one or more channels represent a single modality.  This allows benefiting from the simple diffusion process implementation.

\begin{figure}[h!]
\centering
\includegraphics[width=\linewidth]{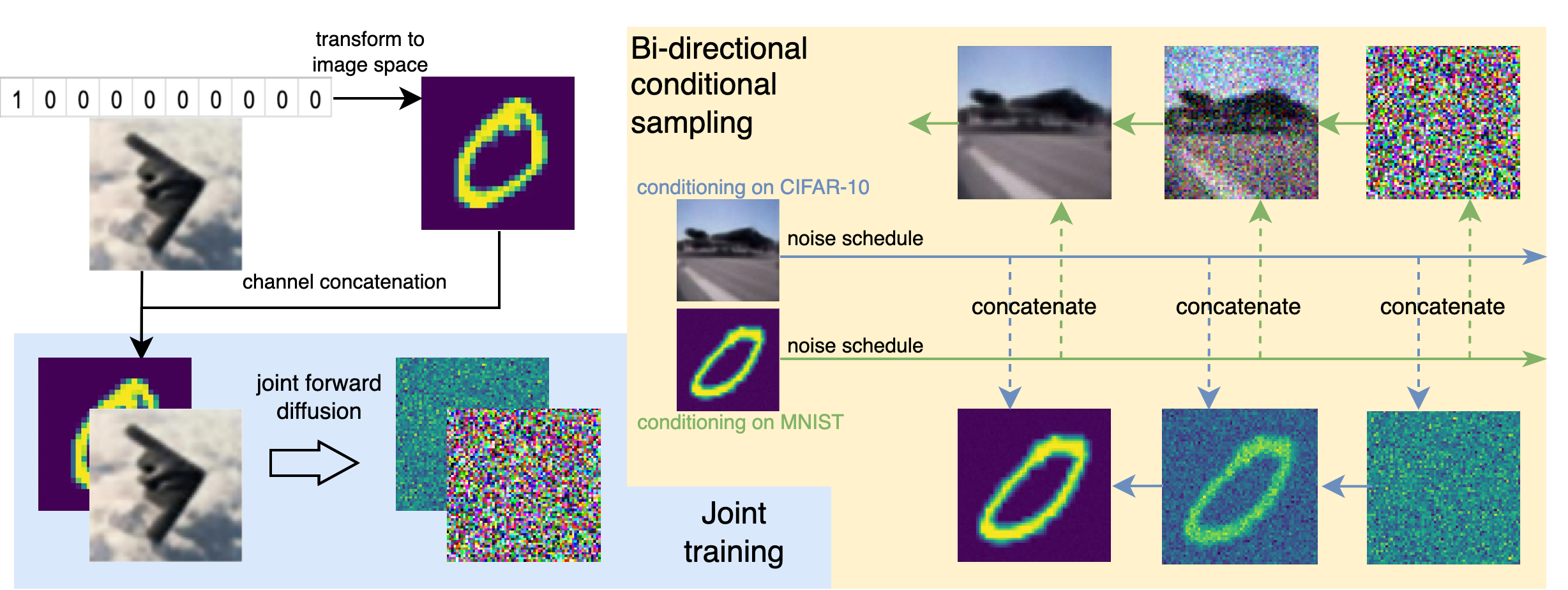}
\caption{The training and sampling pipeline for the proposed multimodal channel-wise image-guided diffusion training: Multimodal data is  transformed into image space via a concatenated channel-wise scheme. The joint diffusion process is learned through a neural network that acquires correlations and associations between the modalities during training. The reverse sampling process can then be conditioned by partial channel guidance, enabling cross-modal conditional generation.}
\label{fig:pipe}
\end{figure}

A multi-channel 2-dimensional array can be a good structure to transform many possible modalities into a shared input space. When passed through the UNet architecture~\cite{ronneberger2015u}, the local $2D$-spatial relationships can be learned, which is ubiquitous in natural data received by humans. This formulation is analogous to the fact that the outermost localized receptors receive pressure and light signals similar to this form of input. The channel-wise correlation is weakly learned compared to the height-width space which allows building the connections between modalities. At the same time, we can still maintain some level of discrepancy between the modalities.
Thus, we can modify denoising diffusion probabilistic models (DDPM) training~\cite{ho2020denoising} and conditional sampling schemes following these ideas, presented in Algorithm 1-3. Algorithms 1 and 2 present our main idea which is pipelined in Figure \ref{fig:pipe}. As for conditional sampling, since at each reverse sampling step, we follow the direction $\nabla_{(x, x')} \log p(x, x')$,   $x'_t$ can be used to provide conditions by using the conditioning image (channel(s)) while maintaining a similar noise level seen in training at timestamp $t$. We proposed two methods to achieve this ability by continuously modifying the guiding channel to a noise-scheduled conditioning image throughout the timestamps. Algorithm 2 follows a random noise schedule consistent with the training scheme of diffusion processes. Finally, Algorithm 3 follows the predicted noise schedule. Details about the mathematical derivation of our algorithms are included in the Appendix.

\begin{algorithm}
\footnotesize
\caption{Channel-wise Image-Guided Diffusion Training}
\label{alg:diffusion_training}
\begin{algorithmic}[1]
\Require{}

\Repeat
\State $(x, x',x''...) \sim p(x, x', x''...)$ \Comment{Sample all correlated data modalities}
\State channel-wise concatenation of $(x, x',x''...)$ to $x^*$
\State training denoising diffusion model $f_{\theta}$ on input $x^*$
\Until{converged}
\end{algorithmic}
\end{algorithm}

\begin{algorithm}[H]
\footnotesize
\caption{Channel-wise Image-Guided Diffusion Sampling with Random Noise Schedule}
\begin{algorithmic}[1]

\Require Target generation noisy channel(s) $c_T$, guiding channel(s) $c'_T$, noise schedule $\{\beta_t\}_{t=T}^{0}$, learned model $f_{\theta}$
\State $c_T, c^{'}_{T} \sim \mathcal{N}(0, I)$
\For{$t = T$ to $1$}
\State Sample two tensors of Gaussian noise $\epsilon_t^{c_t}, \epsilon_t^{c'_t} \sim \mathcal{N}(0, I)$
\If{$t < T$} 
\State Update the guiding channel(s) by $c'_{t} = \sqrt{1 - \beta_{t-1}} \cdot c'_{t-1} + \sqrt{\beta_{t-1}} \cdot \epsilon_t^{c'_{t-1}}$
\EndIf
\State Concatenate the generation channel(s) $c_t$ and the guiding channel(s) $c'_t$ into a single image input $z_t = [c_t, c'_t]$

\State Update the generation channel(s) by $c_{t-1} = \frac{1}{\sqrt{1 - \beta_t}} \cdot c_t - \frac{\beta_t}{\sqrt{1 - \bar{\beta}_t}} \cdot f_{\theta}(z_t, t) + \sqrt{\beta_t} \cdot \epsilon_t^{c_t}$

\EndFor

\Ensure Final denoised image channel(s) $c_0$

\end{algorithmic}
\end{algorithm}

\begin{algorithm}[H]
\footnotesize
\caption{Channel-wise Image-Guided Diffusion Sampling with Predicted Noise Schedule}
\begin{algorithmic}[1]

\Require Target generation noisy channel(s) $c_T$, guiding channel(s) $c'_T$, noise schedule $\{\beta_t\}_{t=T}^{0}$, learned model $f_{\theta}$
\State $c_T, c^{'}_{T} \sim \mathcal{N}(0, I)$
\For{$t = T$ to $1$}
\State Sample only one tensor of Gaussian noise $\epsilon_t^{c_t},  \sim \mathcal{N}(0, I)$
\State Concatenate the generation channel(s) $c_t$ and the guiding channel(s) $c'_t$ into a single image input $z_t = [c_t, c'_t]$
\State Calculate $f_{\theta}(z_t, t)$
\State Update the guiding channel(s) by $c'_{t-1} = \sqrt{1 - \beta_{t-2}} \cdot c'_{t-2} + \sqrt{\beta_{t-2}} \cdot f_{\theta}(z_t, t)$

\State Update the generation channel(s) by $c_{t-1} = \frac{1}{\sqrt{1 - \beta_t}} \cdot c_t - \frac{\beta_t}{\sqrt{1 - \bar{\beta}_t}} \cdot f_{\theta}(z_t, t) + \sqrt{\beta_t} \cdot \epsilon_t^{c_t}$

\EndFor

\Ensure Final denoised image channel(s) $c_0$

\end{algorithmic}
\end{algorithm}

\section{Experimental Validation}

We validate our approach empirically. Our code is available as a supplement.

\subsection{Experimental Setup}

\subsubsection{Datasets}
Given that prior benchmarks do not exist for our purpose, we generate data with concurrent modalities by concatenating two labeled datasets and aligning them during training. We demonstrate that our approach is able to generate cross-modal data in a conditional manner. Although our multimodal dataset is not a real-life dataset, but it models associations between modalities relatively well. We use the MNIST and the CIFAR-10 datasets as one modality.  We used these datasets to match the computational resources we have. The reason for selecting the MINST dataset is to simulate written language such that it is correlated with the identity of the CIFAR-10 objects. In other words, if our approach works well, we can transform the language modality, i.e., MNIST, to the image space, i.e., CIFAR-10. We think that other modalities can be transformed into this space given the right modeling of their natural input pathways as a 2-D spatial format. For the moment, we generate results for two modalities.
We increase the resolution of both datasets for adding more details which enable better visual evaluation. Extra information on CIFAR-10 is also added through an upscaling super-resolution model.
During joint training, 5000 samples from each class in CIFAR-10 are selected to concatenate with samples from MNIST with matching numerical classes, i.e., we assume that classes with the same ID correlate across the two modalities. MNIST samples are selected with repetition.

\subsubsection{Baselines}
 We compared our results from different image sampling approaches to demonstrate our method is robust. We use (i) Unconditional joint generation, (ii) conditional generation without noise scheduling (a constant image from the test set is used throughout the steps of the reverse process), (iii) conditional generation with random noise scheduling (a random noise is used to corrupt the conditioning image at each timestamp), and (iv) conditional generation with predicted noise scheduling (the predicted noise from timestamp $t$ is used to corrupt the conditioning image at timestamp $t-1$). The experiments are conducted using a single jointly trained UNet architecture.
\subsection{Evaluation Methodology}
We use FID score~\cite{heusel2017gans} and IS~\cite{salimans2016improved} which are common metrics in generative AI. We also report the average precision and recall of ten classes to evaluate the class separation of the generated samples. For FID, we provide random Gaussian noise-generated inception logits compared with the inception logits generated by CIFAR-10 and MNIST test set as the upper-bound. The same noise is used for IS as a lower bound. For precision and recall, we use two separately trained CNNs and provide the test set precision/recall as the upper bound performance for CIFAR-10 and MNIST. 
For the unconditional joint generation without true labels, we evaluate the precision and recall   using the method of counting matching prediction results by setting the prediction from one modality to the other.

\subsection{Results}

\subsubsection{Unconditional Joint Generation}
\begin{figure}[h!]
    \centering
        \centering
        \includegraphics[width = \linewidth]{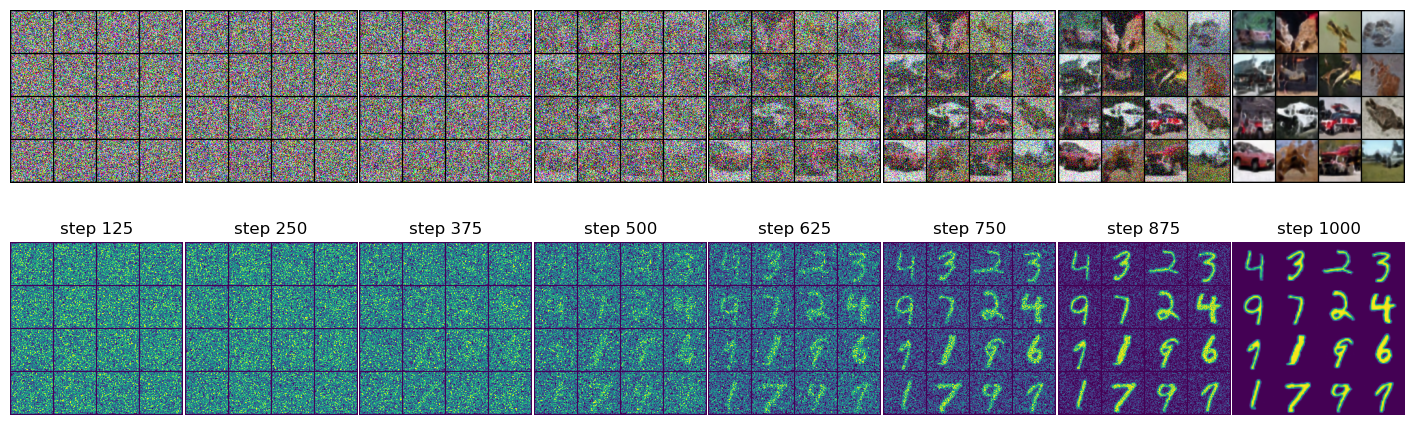}
        \caption{Unconditional Generation Generates Paired Images}
       \label{fig:uncon}
\end{figure}

\begin{figure}[h!]
    \centering
    \begin{subfigure}[b]{0.49\textwidth}
        \centering
        \includegraphics[width = \linewidth]{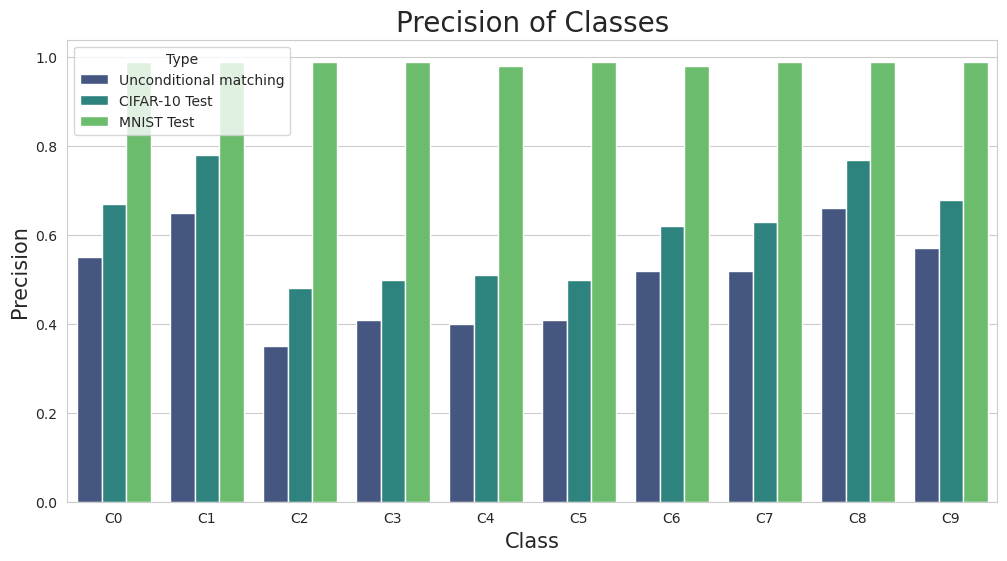}
        \caption{CIFAR10 generation conditioned on MNIST}
        \label{figbar:subfig1}
    \end{subfigure}
    \hfill
    \begin{subfigure}[b]{0.49\textwidth}
        \centering
        \includegraphics[width = \linewidth]{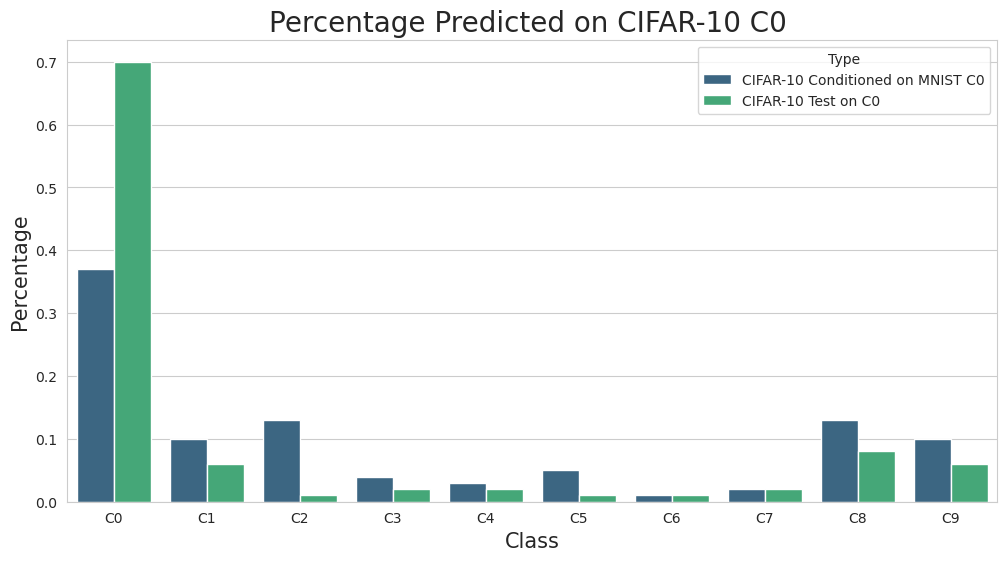}
        \caption{MNIST generation conditioned on CIFAR-10}
        \label{figbar:subfig2}
    \end{subfigure}
    \caption{Comparison between joint generation prediction precision with a classifier. The classifier is trained on the original MNIST or CIFAR-10 training data, and evaluated on the test data. Unconditional matching is the matching classifier prediction outputs method we use to evaluate the precision of the jointly generated image.}
    \label{fig:bar}
\end{figure}

We first tested the unconditional joint sampling of all 4 channels without any guidance. Figure \ref{fig:uncon} presents our results for this purpose. We observe that our approach is able to generate visually correlated classes from both CIFAR-10 and MNIST concurrently which as we will see enables conditional data generation. This ability is also acquired during the training phase. To evaluate the quality of the generated samples, Table \ref{table:comparison} presents our comparison. We can see that our approach generates images that  have the best quality in general for both modalities.  Visually, the MNIST samples are more recognizable due to the more distinct features. We also see in Figure \ref{figbar:subfig1} that when matching the two classifiers' performance on the jointly generated samples, the class-wise matching precision has a high correlation with the CIFAR-10 classifier test on the CIFAR-10 test set, showing that the lower-variance MNIST condition provides some guidance that is similar to a classifier.

\begin{figure}[h!]
    \centering
    \begin{subfigure}[b]{0.44\textwidth}
        \centering
        \includegraphics[width = \linewidth]{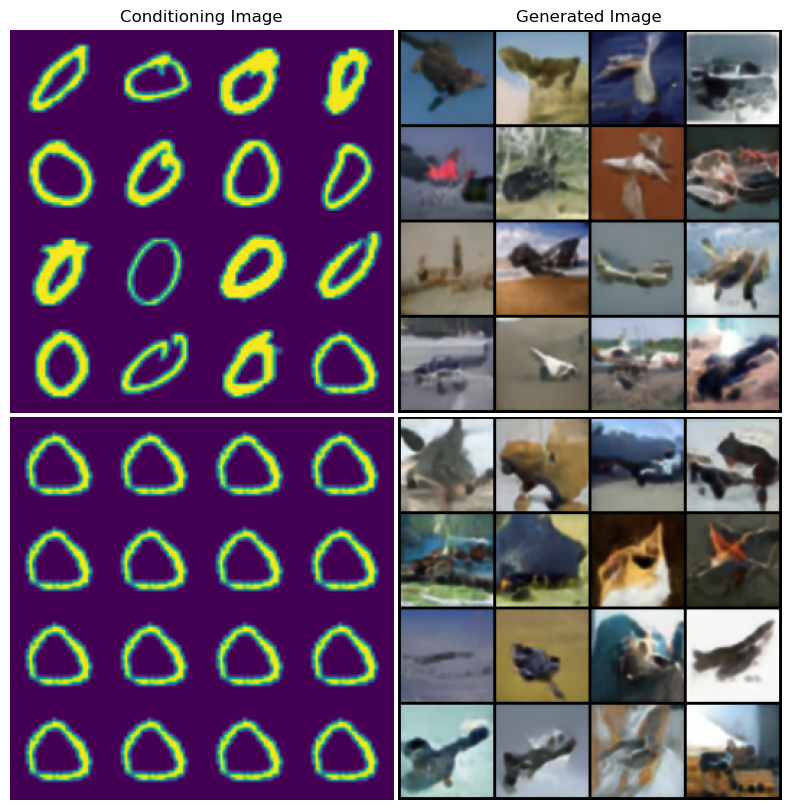}
        \caption{CIFAR-10 generation conditioned on MNIST}
        \label{figcon_r:subfig1}
    \end{subfigure}
    \hfill
    \begin{subfigure}[b]{0.44\textwidth}
        \centering
        \includegraphics[width = \linewidth]{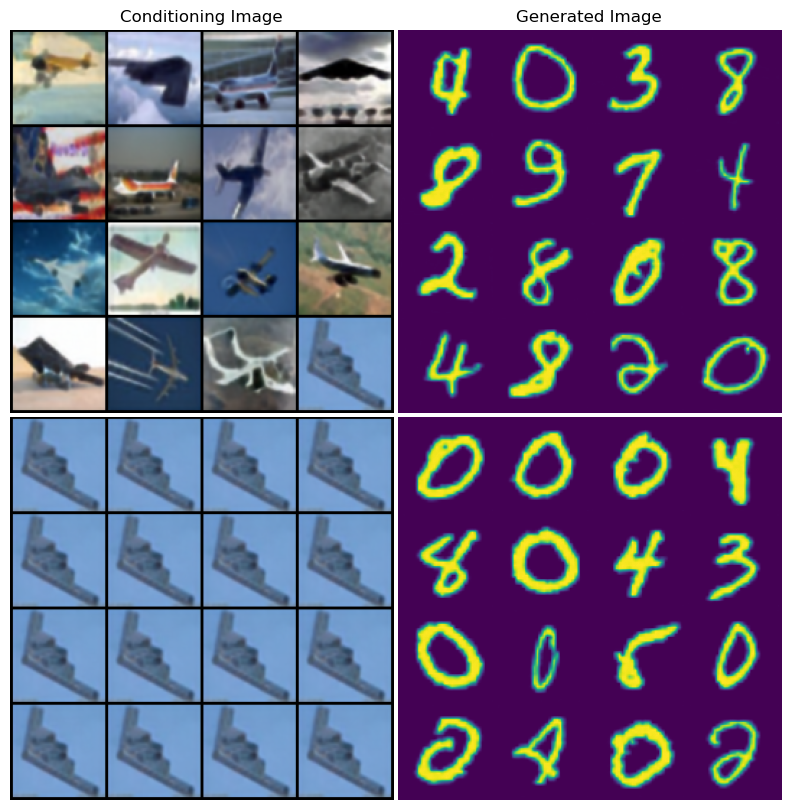}
        \caption{MNIST generation conditioned on CIFAR-10}
        \label{figcon_r:subfig2}
    \end{subfigure}
    \caption{Results for channel-wise image-guided generation, with random noise schedule applied on the conditioning image: (top) random instances in class ``0''; (bottom) fixed instance in class ``0'')}
    \label{fig:con_r}
\end{figure}

\subsubsection{Conditional Generation}
\begin{figure}[h!]
    \centering
    \begin{subfigure}[b]{0.44\textwidth}
        \centering
        \includegraphics[width = \linewidth]{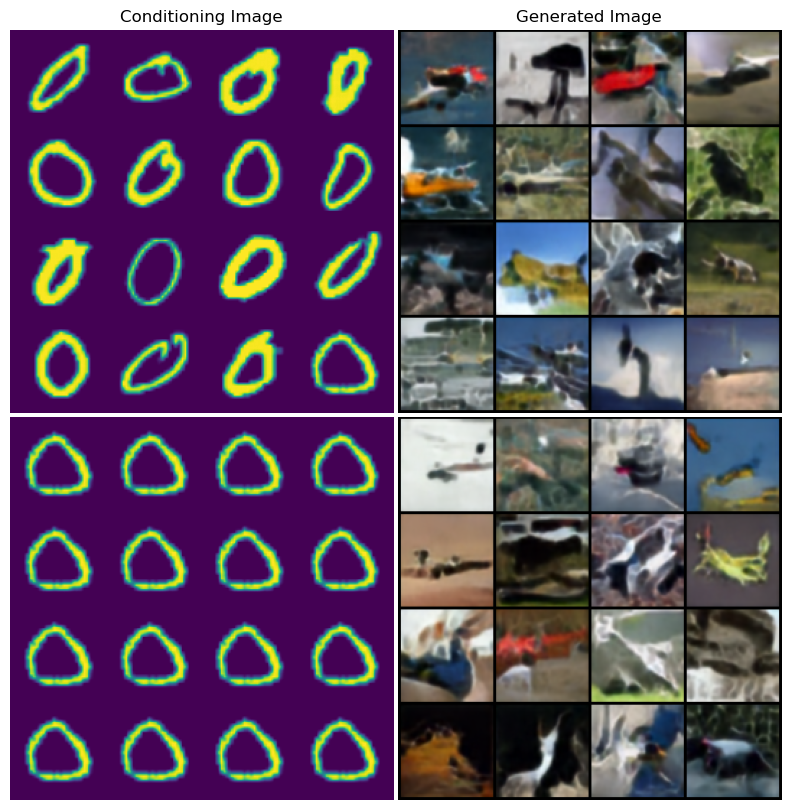}
        \caption{CIFAR10 generation conditioned on MNIST}
        \label{figcon_p:subfig1}
    \end{subfigure}
    \hfill
    \begin{subfigure}[b]{0.44\textwidth}
        \centering
        \includegraphics[width = \linewidth]{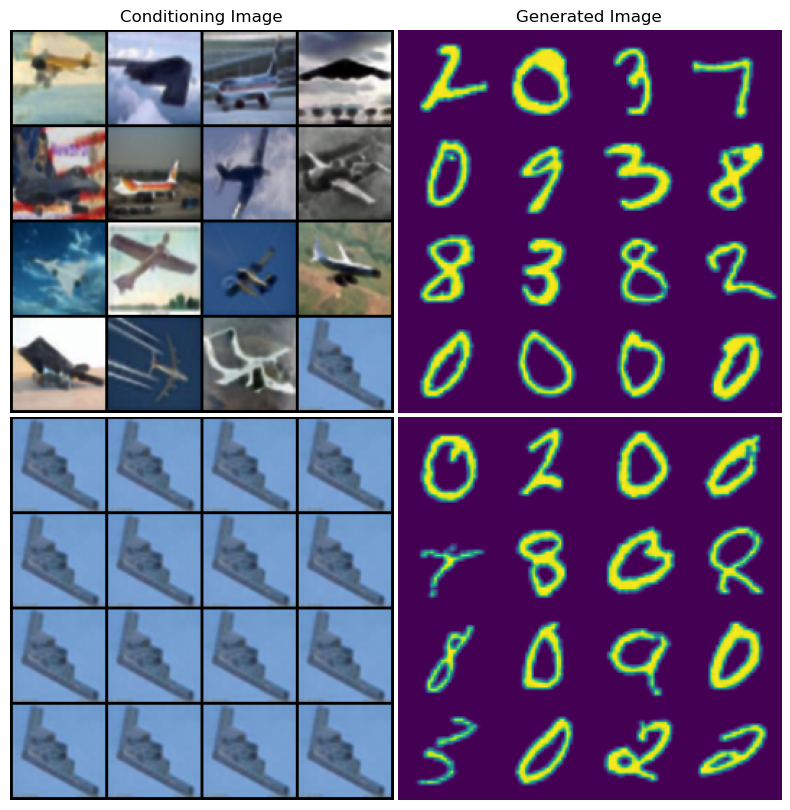}
        \caption{MNIST generation conditioned on CIFAR-10}
        \label{figcon_p:subfig2}
    \end{subfigure}
    \caption{Channel-wise image-guided generation, with predicted noise schedule applied on the conditioning image. (Top: random instances in class ``0''; Bottom: fixed instance in class ``0'')}
    \label{fig:con_p}
\end{figure}

\begin{figure}[h!]
    \centering
    \begin{subfigure}[b]{0.44\textwidth}
        \centering
        \includegraphics[width = \linewidth]{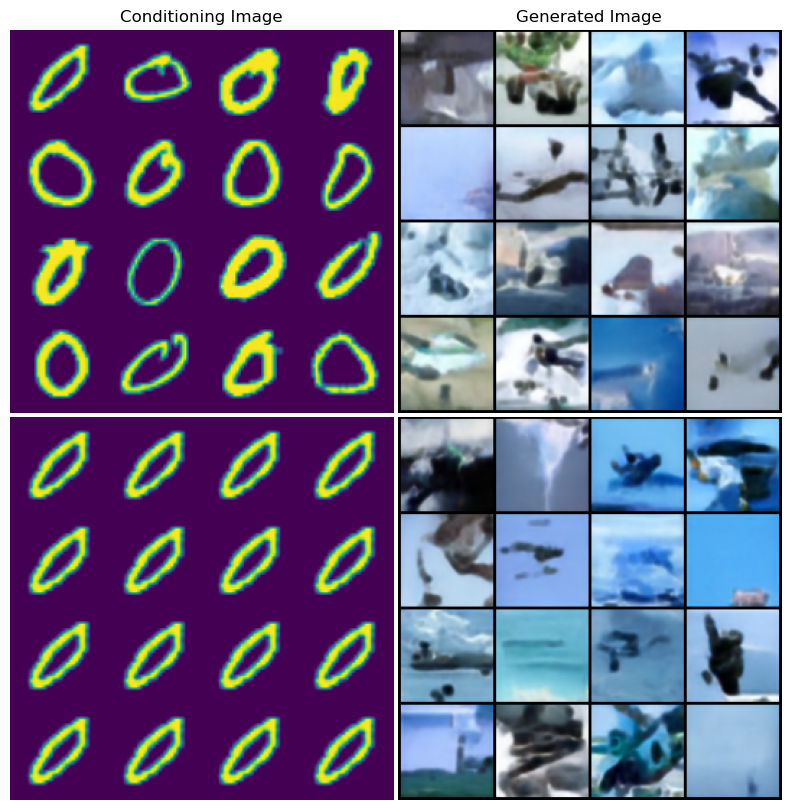}
        \caption{CIFAR10 generation conditioned on MNIST}
        \label{figcon_n:subfig1}
    \end{subfigure}
    \hfill
    \begin{subfigure}[b]{0.44\textwidth}
        \centering
        \includegraphics[width = \linewidth]{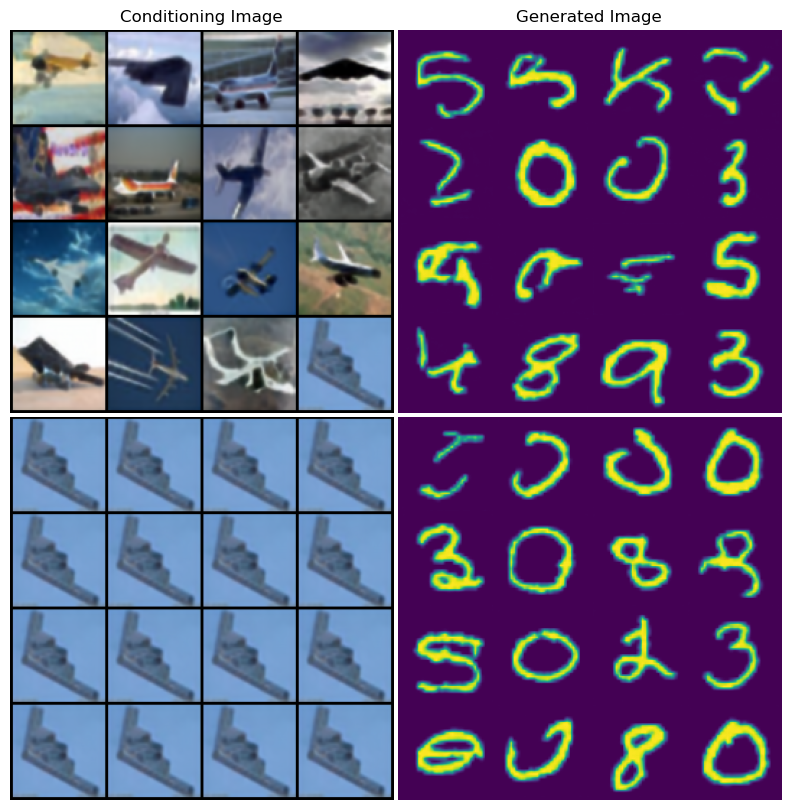}
        \caption{MNIST generation conditioned on CIFAR-10}
        \label{figcon_n:subfig2}
    \end{subfigure}
    \caption{Channel-wise image-guided generation, with a constant guiding image with no noise. (Top: random instances in class ``0''; Bottom: fixed instance in class ``0'')}
    \label{fig:con_n}
\end{figure}

Figure \ref{figcon_r:subfig1} presents the generation of CIFAR-10 images conditioned on MNIST images, with random noise scheduling on the guiding image. Figure \ref{figcon_p:subfig1} and Figure \ref{figcon_n:subfig1} present the generation with predicted noise and no noise. This figure exhibits the capability of relating  MNIST class ``0'' to the CIFAR-10 class ``airplane''. We further examined the predicted classes on CIFAR-10 generated images conditioned on MNIST class ``0'', i.e., corresponding to CIFAR-10 ``airplane'' during training. In Figure \ref{figbar:subfig2}, we also observe a high correlation with the CIFAR-10 classifier on the CIFAR-10 test set. In addition, classes such as C2'bird', C8'ship', C1'automobile', and C9'truck' are the top predictions, which are either man-made machines or flying objects sharing some key features with ``airplane''. This observation suggests that the channel-wise image guidance also helps learning more meaningful latent space that encodes some shared features among these classes. 
This is a significant observation because it shows our approach can lead to training latent spaces that are semantically meaningful. Such a space can be used for transfer learning, e.g., zero-shot learning or domain adaptation settings.

For the other direction of generation, where we generate MNIST images conditioned on CIFAR-10 image (Figure \ref{figcon_r:subfig2},\ref
{figcon_p:subfig2},\ref{figcon_n:subfig2}), the average precision dropped, mainly due to the randomness which happens at the earlier reverse steps. This randomness impacts  CIFAR-10 images more than the MNIST images due to the complexity differences when the MNIST image settles to a class before CIFAR-10. However, we observe in Figure \ref{figcon_n:subfig2}  that when we run the generation step multiple times on a single conditioning image, the majority of generated MNIST images still have a matching class.

When comparing the three different image-guiding methods, the constant no-noise guiding fails to capture the detailed features of a class and generates relatively less-variant biased outputs. The random noise scheduling approach generates the best-quality images while performing worse than the predicted noise scheduling. This is likely due to the predicted noise providing more guidance from the conditioning image, trading off the unconditional generation quality, similar to a higher weight on the classifier score function in the classifier-guided diffusion model.

\begin{table}[h!]
\centering
\footnotesize	

\begin{tabular}{clcccc}
\toprule
 Generation & Setting & FID & IS & Precision & Recall \\
\midrule
\multirow{4}{*}{CIFAR-10} & UPPER BOUND & 3013.4 & - & 0.61 & 0.61 \\
 & unconditional joint & 194.94 & 4.170 & 0.50* & 0.51*  \\
 & conditioned on MNIST w random noise & \textbf{215.2} & \textbf{3.919} & 0.36 & 0.36 \\
  & conditioned on MNIST w predicted noise & 222.0 & 3.845 & \textbf{0.42} & \textbf{0.41}\\
 & conditioned on MNIST w/o noise & 330.82 & 3.39 & 0.26 & 0.21 \\
 & LOWER BOUND & - & 1.040 & 0.09 & 0.09 \\
\midrule
\multirow{4}{*}{MNIST} & UPPER BOUND & 2811.54 & - & 0.99 & 0.99 \\
 & unconditional joint & 15.28 & 2.314 & 0.50* & 0.51*  \\
 & conditioned on CIFAR-10 w random noise & \textbf{16.48} & 2.297 & 0.21 & 0.21\\
 & conditioned on CIFAR-10 w predicted noise & 20.86 & \textbf{2.360} & \textbf{0.26} & \textbf{0.26} \\
 & conditioned on CIFAR-10 w/o noise & 40.18 & 2.203 & 0.22 & 0.20 \\
 & LOWER BOUND & - & 1.040 & 0.09 & 0.09 \\
\bottomrule
\end{tabular}
\caption{Comparison of generation quality and classification performance using four metrics. * is a pseudo evaluation by counting the matching predicted classes of the two jointly generated images. All the three conditional generation methods showed the effectiveness of the joint training. Numbers in bold denote the best conditional generation performance in each column.}
\label{table:comparison}

\end{table}

\section{Conclusions}
We introduced a novel approach for cross-modal data generation using a channel-wise image-guided diffusion model. Our method effectively generates data across different modalities in all directions given only noisy input and showcases its potential in understanding inter-modality relationships in a real-life setting.
The model's performance, demonstrated on the channel-wise concatenated CIFAR-10 and MNIST datasets, suggests a robust framework for combining various data modalities into a single modality in the input space. The results also indicate the approach's capability in learning meaningful latent spaces and shared features among classes.
Our work validates that a cognitively inspired cross-model generation method is effective using a  single model. We did not use more complex datasets in our experiments to match our computational resources but we hope this work can provide an initial insight to approach cross-modal generation in a different but intriguing direction. 
Looking ahead, our focus will be on enhancing the robustness and flexibility of our model. This process includes extending it to more complex modalities, and refining noise scheduling strategies for each modality separately. Moreover, we plan to explore practical applications like cross-modal data understanding with real-time multi-modal data, fast noisy input conditioned generation, and so on.

{
\small
\bibliography{reference}{}

\begin{thebibliography}{10}

\bibitem{hipo}
P.~J. Brasted, T.~J. Bussey, E.~A. Murray, and S.~P. Wise.
\newblock {Role of the hippocampal system in associative learning beyond the
  spatial domain}.
\newblock {\em Brain}, 126(5):1202--1223, 05 2003.

\bibitem{Classifier}
Prafulla Dhariwal and Alex Nichol.
\newblock Diffusion models beat gans on image synthesis.
\newblock {\em CoRR}, abs/2105.05233, 2021.

\bibitem{dhariwal2021diffusion}
Prafulla Dhariwal and Alexander Nichol.
\newblock Diffusion models beat gans on image synthesis.
\newblock {\em Advances in Neural Information Processing Systems},
  34:8780--8794, 2021.

\bibitem{gan}
Ian~J. Goodfellow, Jean Pouget-Abadie, Mehdi Mirza, Bing Xu, David
  Warde-Farley, Sherjil Ozair, Aaron Courville, and Yoshua Bengio.
\newblock Generative adversarial networks, 2014.

\bibitem{hebb}
Donald Hebb.
\newblock The organization of behavior. emphnew york, 1949.

\bibitem{heusel2017gans}
Martin Heusel, Hubert Ramsauer, Thomas Unterthiner, Bernhard Nessler, and Sepp
  Hochreiter.
\newblock Gans trained by a two time-scale update rule converge to a local nash
  equilibrium.
\newblock {\em Advances in neural information processing systems}, 30, 2017.

\bibitem{DDPM}
Jonathan Ho, Ajay Jain, and Pieter Abbeel.
\newblock Denoising diffusion probabilistic models.
\newblock {\em CoRR}, abs/2006.11239, 2020.

\bibitem{ho2020denoising}
Jonathan Ho, Ajay Jain, and Pieter Abbeel.
\newblock Denoising diffusion probabilistic models.
\newblock {\em Advances in Neural Information Processing Systems},
  33:6840--6851, 2020.

\bibitem{ho2021classifier}
Jonathan Ho and Tim Salimans.
\newblock Classifier-free diffusion guidance.
\newblock In {\em NeurIPS 2021 Workshop on Deep Generative Models and
  Downstream Applications}.

\bibitem{Classifierfree}
Jonathan Ho and Tim Salimans.
\newblock Classifier-free diffusion guidance, 2022.

\bibitem{hu2023encoding}
Zizhao Hu and Mohammad Rostami.
\newblock Encoding binary concepts in the latent space of generative models for
  enhancing data representation.
\newblock {\em arXiv preprint arXiv:2303.12255}, 2023.

\bibitem{hime}
Joshua~P. Johansen, Lorenzo Diaz-Mataix, Hiroki Hamanaka, Takaaki Ozawa, Edgar
  Ycu, Jenny Koivumaa, Ashwani Kumar, Mian Hou, Karl Deisseroth, Edward~S.
  Boyden, and Joseph~E. LeDoux.
\newblock Hebbian and neuromodulatory mechanisms interact to trigger
  associative memory formation.
\newblock {\em Proceedings of the National Academy of Sciences},
  111(51):E5584--E5592, 2014.

\bibitem{hihe}
S~R Kelso, A~H Ganong, and T~H Brown.
\newblock Hebbian synapses in hippocampus.
\newblock {\em Proceedings of the National Academy of Sciences},
  83(14):5326--5330, 1986.

\bibitem{khattar2019mvae}
Dhruv Khattar, Jaipal~Singh Goud, Manish Gupta, and Vasudeva Varma.
\newblock Mvae: Multimodal variational autoencoder for fake news detection.
\newblock In {\em The world wide web conference}, pages 2915--2921, 2019.

\bibitem{vae}
Diederik~P Kingma and Max Welling.
\newblock Auto-encoding variational bayes, 2022.

\bibitem{pavl}
Ivan~Petrovich Pavlov.
\newblock {\em Conditioned Reflexes: An Investigation of the Physiological
  Activity of the Cerebral Cortex}.
\newblock Oxford University Press, 1927.

\bibitem{pfeiffer2020content}
Brad~E Pfeiffer.
\newblock The content of hippocampal “replay”.
\newblock {\em Hippocampus}, 30(1):6--18, 2020.

\bibitem{hila}
Vitória Piai, Kristopher~L. Anderson, Jack~J. Lin, Callum Dewar, Josef
  Parvizi, Nina~F. Dronkers, and Robert~T. Knight.
\newblock Direct brain recordings reveal hippocampal rhythm underpinnings of
  language processing.
\newblock {\em Proceedings of the National Academy of Sciences},
  113(40):11366--11371, 2016.

\bibitem{DALLE}
Aditya Ramesh, Prafulla Dhariwal, Alex Nichol, Casey Chu, and Mark Chen.
\newblock Hierarchical text-conditional image generation with clip latents,
  2022.

\bibitem{ronneberger2015u}
Olaf Ronneberger, Philipp Fischer, and Thomas Brox.
\newblock U-net: Convolutional networks for biomedical image segmentation.
\newblock In {\em Medical Image Computing and Computer-Assisted
  Intervention--MICCAI 2015: 18th International Conference, Munich, Germany,
  October 5-9, 2015, Proceedings, Part III 18}, pages 234--241. Springer, 2015.

\bibitem{rostami2023overcome}
Mohammad Rostami and Aram Galstyan.
\newblock Overcoming concept shift in domain-aware settings through
  consolidated internal distributions.
\newblock In {\em AAAI Conference on Artificial Intelligence}, 2023.

\bibitem{rostami2020generative}
Mohammad Rostami, Soheil Kolouri, Praveen Pilly, and James McClelland.
\newblock Generative continual concept learning.
\newblock In {\em Proceedings of the AAAI conference on artificial
  intelligence}, volume~34, pages 5545--5552, 2020.

\bibitem{Imagen}
Chitwan Saharia, William Chan, Saurabh Saxena, Lala Li, Jay Whang, Emily
  Denton, Seyed Kamyar~Seyed Ghasemipour, Burcu~Karagol Ayan, S.~Sara Mahdavi,
  Rapha~Gontijo Lopes, Tim Salimans, Jonathan Ho, David~J Fleet, and Mohammad
  Norouzi.
\newblock Photorealistic text-to-image diffusion models with deep language
  understanding, 2022.

\bibitem{salimans2016improved}
Tim Salimans, Ian Goodfellow, Wojciech Zaremba, Vicki Cheung, Alec Radford, and
  Xi~Chen.
\newblock Improved techniques for training gans.
\newblock {\em Advances in neural information processing systems}, 29, 2016.

\bibitem{Diffusion}
Jascha Sohl{-}Dickstein, Eric~A. Weiss, Niru Maheswaranathan, and Surya
  Ganguli.
\newblock Deep unsupervised learning using nonequilibrium thermodynamics.
\newblock {\em CoRR}, abs/1503.03585, 2015.

\bibitem{SMLD}
Yang Song and Stefano Ermon.
\newblock Generative modeling by estimating gradients of the data distribution.
\newblock {\em CoRR}, abs/1907.05600, 2019.

\bibitem{triche2022exploration}
Anthony Triche, Anthony~S Maida, and Ashok Kumar.
\newblock Exploration in neo-hebbian reinforcement learning: Computational
  approaches to the exploration-exploitation balance with bio-inspired neural
  networks.
\newblock {\em Neural Networks}, 2022.

\end{thebibliography}
\bibliographystyle{plain}
}




\newtheorem{theorem}{Theorem}
\newenvironment{proof}{\paragraph{Proof:}}{\hfill$\square$}

\clearpage 

\appendix 
\section*{Supplementary Material}

\section{Additional Experiments}
Here, we show the additional experiment results that are implemented on different class labels. Each class label, ranging from 0 to 9, represents a unique category. Specifically, \textbf{Class 0} is for images of \textbf{airplanes}, \textbf{Class 1} is for \textbf{automobiles}, \textbf{Class 2} is for \textbf{birds}, \textbf{Class 3} is for \textbf{cats}, \textbf{Class 4} is for \textbf{deer}, \textbf{Class 5} is for \textbf{dogs}, \textbf{Class 6} is for \textbf{frogs}, \textbf{Class 7} is for \textbf{horses}, \textbf{Class 8} is for \textbf{ships}, and finally, \textbf{Class 9} is for images of \textbf{trucks}.

\begin{figure}[htbp]
    \centering
    \caption{Generated MNIST image conditioned on CIFAR-10 image}
    \subfloat[]{\includegraphics[width=0.45\textwidth, keepaspectratio]{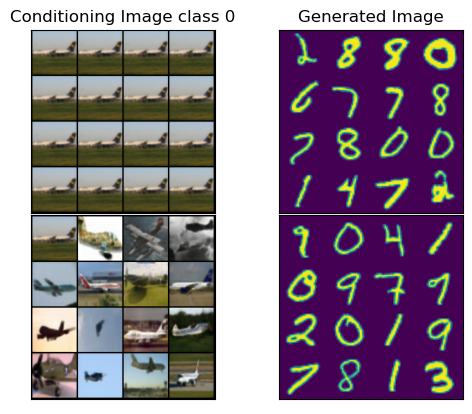}\label{fig:mnist0}}
    \hfill
    \subfloat[]{\includegraphics[width=0.45\textwidth, keepaspectratio]{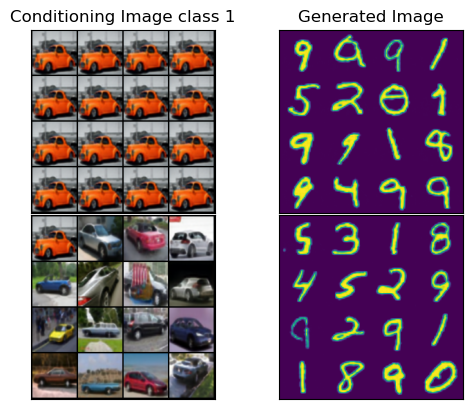}\label{fig:mnist1}}
    
    \subfloat[]{\includegraphics[width=0.45\textwidth, keepaspectratio]{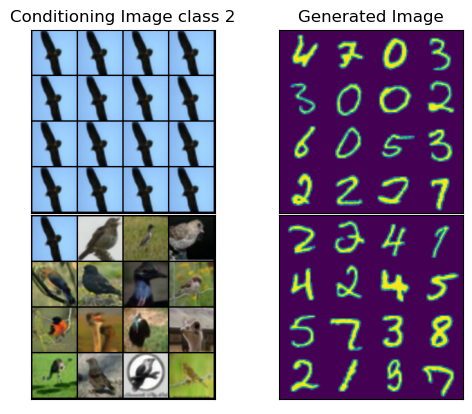}\label{fig:mnist2}}
    \hfill
    \subfloat[]{\includegraphics[width=0.45\textwidth, keepaspectratio]{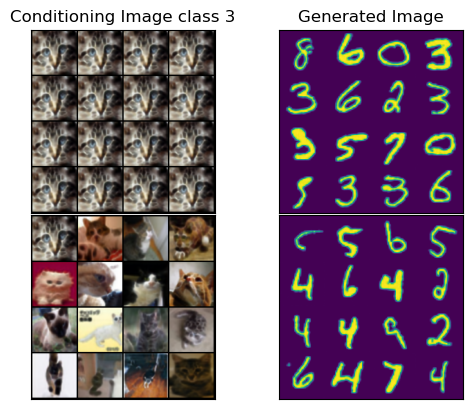}\label{fig:mnist3}}

\end{figure}

\clearpage

\begin{figure}[htbp]
    \ContinuedFloat
    \centering
    \caption{Generated MNIST image conditioned on CIFAR-10 image (Continued)}
    \subfloat[]{\includegraphics[width=0.45\textwidth, keepaspectratio]{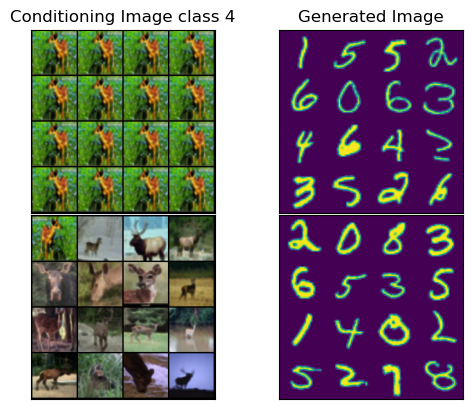}\label{fig:mnist4}}
    \hfill
    \subfloat[]{\includegraphics[width=0.45\textwidth, keepaspectratio]{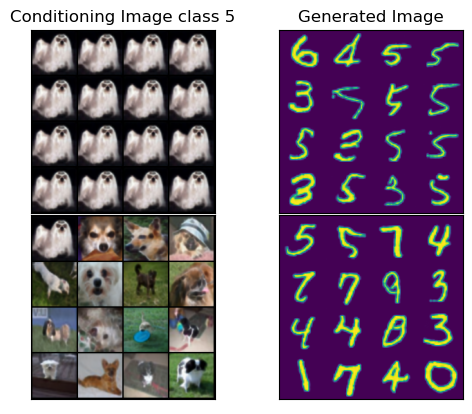}\label{fig:mnist5}}
    
    \subfloat[]{\includegraphics[width=0.45\textwidth, keepaspectratio]{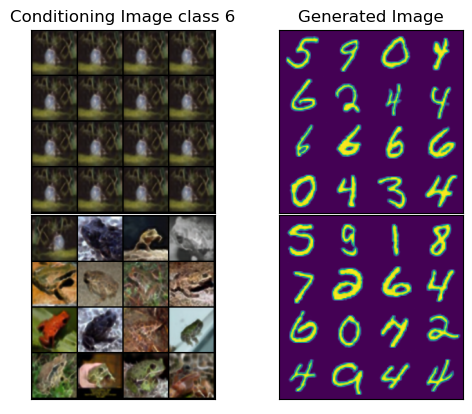}\label{fig:mnist6}}
    \hfill
    \subfloat[]{\includegraphics[width=0.45\textwidth, keepaspectratio]{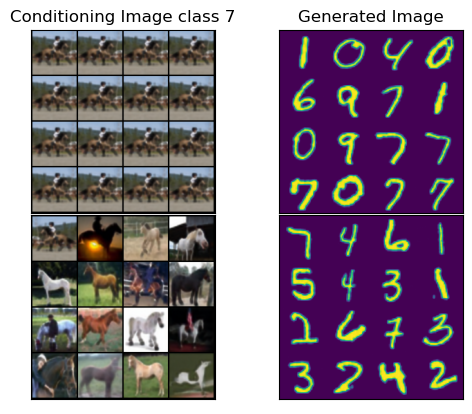}\label{fig:mnist7}}
    
    \subfloat[]{\includegraphics[width=0.45\textwidth, keepaspectratio]{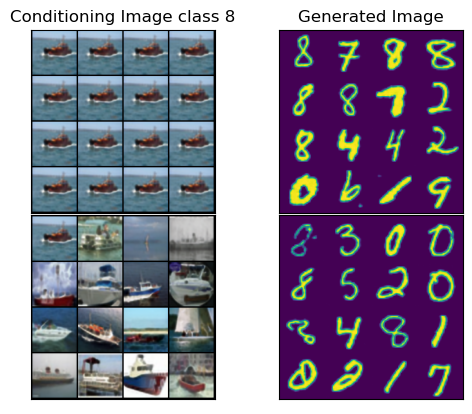}\label{fig:mnist8}}
    \hfill
    \subfloat[]{\includegraphics[width=0.45\textwidth, keepaspectratio]{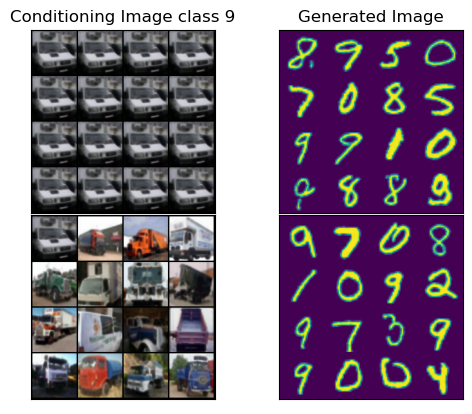}\label{fig:mnist9}}
\end{figure}

\begin{figure}[htbp]
    \centering
    \caption{Generated CIFAR-10 image conditioned on MNIST image}
    \subfloat[]{\includegraphics[width=0.45\textwidth, keepaspectratio]{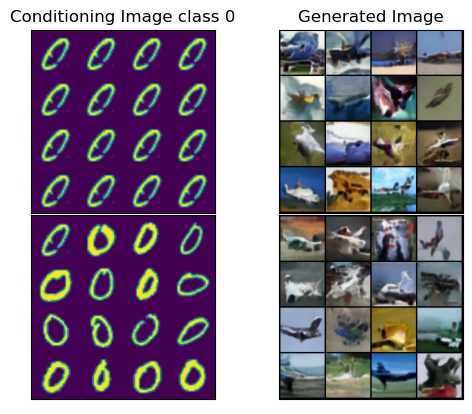}\label{fig:cifar0}}
    \hfill
    \subfloat[]{\includegraphics[width=0.45\textwidth, keepaspectratio]{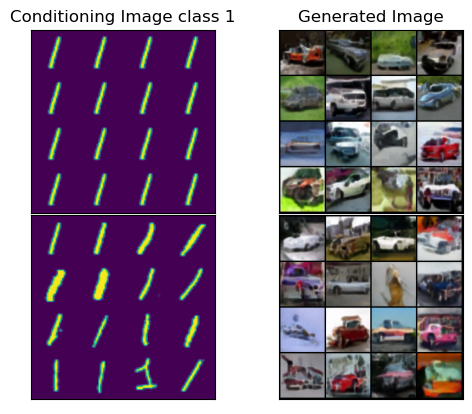}\label{fig:cifar1}}
    
    \subfloat[]{\includegraphics[width=0.45\textwidth, keepaspectratio]{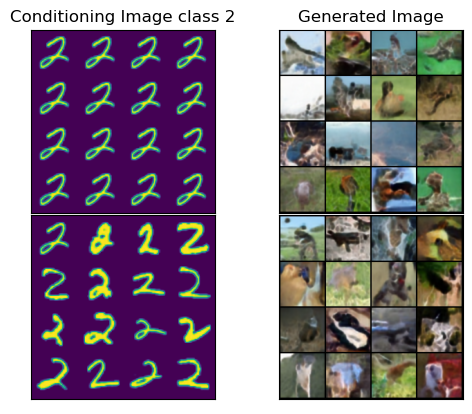}\label{fig:cifar2}}
    \hfill
    \subfloat[]{\includegraphics[width=0.45\textwidth, keepaspectratio]{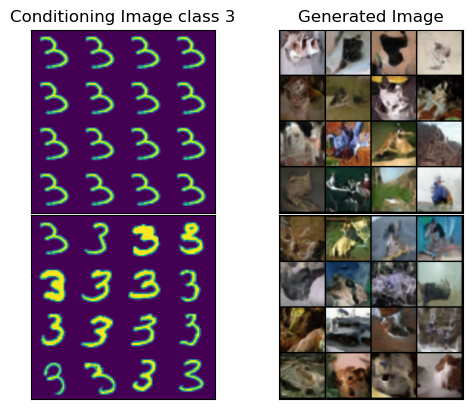}\label{fig:cifar3}}

    \subfloat[]{\includegraphics[width=0.45\textwidth, keepaspectratio]{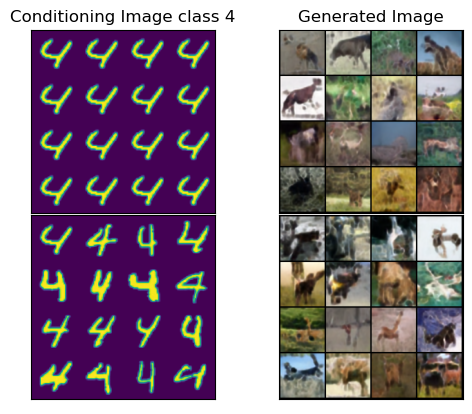}\label{fig:cifar4}}
    \hfill
    \subfloat[]{\includegraphics[width=0.45\textwidth, keepaspectratio]{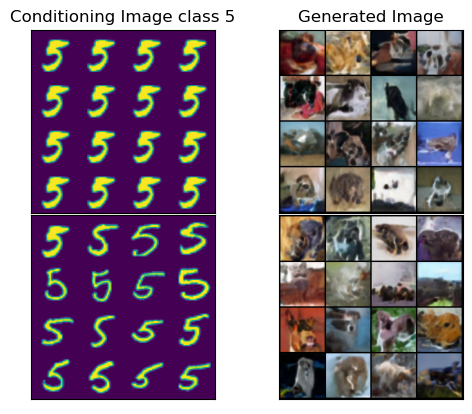}\label{fig:cifar5}}
\end{figure}

\clearpage

\begin{figure}[htbp]
    \ContinuedFloat
    \centering
    \caption{Generated CIFAR-10 image conditioned on MNIST image (Continued)}
    \subfloat[]{\includegraphics[width=0.45\textwidth, keepaspectratio]{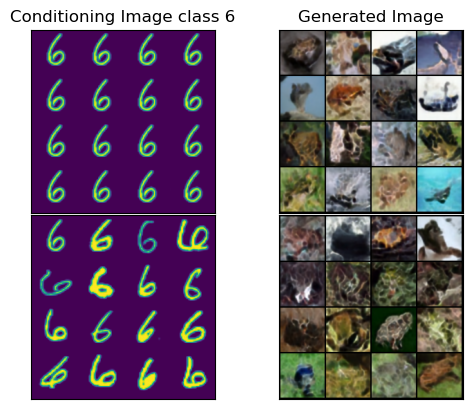}\label{fig:cifar6}}
    \hfill
    \subfloat[]{\includegraphics[width=0.45\textwidth, keepaspectratio]{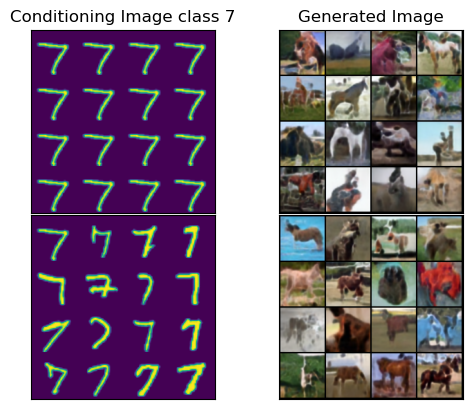}\label{fig:cifar7}}

    \subfloat[]{\includegraphics[width=0.45\textwidth, keepaspectratio]{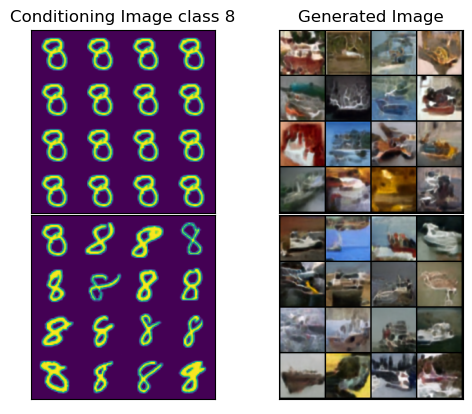}\label{fig:cifar8}}
    \hfill
    \subfloat[]{\includegraphics[width=0.45\textwidth, keepaspectratio]{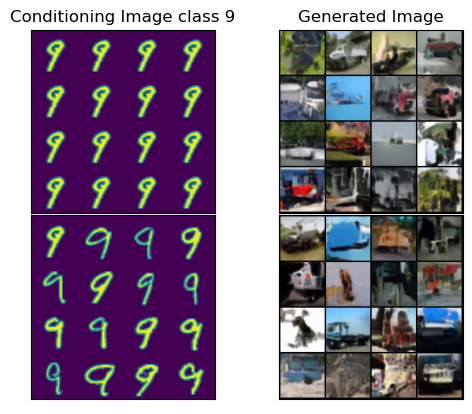}\label{fig:cifar9}}
\end{figure}
\section{Explanation of Algorithms}

\subsection{Algorithm 2}

We present proof of the diffusion sampling algorithm that operates on concatenated images with channel-wise image-guided sampling. Our goal is to show that the algorithm will correctly propagate information from the guiding channels to the generation channels and yield a final denoised image.

\begin{enumerate}
\item \textbf{Initialization (Line 1):} The algorithm first generates the generation and guiding channels using a Gaussian noise:
\begin{equation}
c_T, c'_T \sim \mathcal{N}(0, I)
\end{equation}
This is a standard initial condition for diffusion processes.

\item \textbf{Sampling Gaussian noise (Line 3):} Two tensors of Gaussian noise are sampled for the generation channel and the guiding channel:
\begin{equation}
\epsilon_{ct}^{t}, \epsilon_{c't}^{t} \sim \mathcal{N}(0, I)
\end{equation}

\item \textbf{Updating the Guiding Channel (Lines 4-6):} The guiding channels are updated according to the noise schedule $\{\beta_t\}_{t=0}^{T}$ applied to the guiding image at the current timestamp:
\begin{equation}
c'_t = \sqrt{1 - \beta_{t-1}} \cdot c'_{t-1} + \sqrt{\beta_{t-1}} \cdot \epsilon_{c't}^{t-1}
\end{equation}

\item \textbf{Concatenation (Line 7):} The generation channels and the guiding channels are concatenated into a single image input:
\begin{equation}
z_t = [c_t, c'_t]
\end{equation}
This concatenated input serves as the input for the learned model $f_{\theta}$. And this is where the information propagation happens. The concatenated channels will provide a joint prediction of noise through the model, which provides cross-channel connections. 

\item \textbf{Updating the Generation Channel (Line 8):} The generation channel is updated using the reverse denoising process where the output of the model $f_{\theta}$, the predicted joint noise, is denoised from a noisier version of the generation channel(note: the guiding channel will be updated in the next iteration anyway, so we don't have to denoise the guiding channels here):
\begin{equation}
c_{t-1} = \frac{1}{\sqrt{1 - \beta_t}} \cdot c_t - \frac{\beta_t}{\sqrt{1 - \overline{\beta}_t}} \cdot f_{\theta}(z_t, t) + \sqrt{\beta_t} \cdot \epsilon_{ct}^{t}
\end{equation}
\end{enumerate}

\subsection{Algorithm 3}

Algorithm 3 is similar to Algorithm 2, so we will focus on the differences from Algorithm 2.

\begin{enumerate}
\item \textbf{Sampling Gaussian noise (Line 3):} In contrast to Algorithm 2, Algorithm 3 samples only one tensor of Gaussian noise for the generation channel:
\begin{equation}
\epsilon_{ct}^{t} \sim \mathcal{N}(0, I)
\end{equation}

    \item \textbf{Prediction with $f_{\theta}$ (Line 5):} Unlike Algorithm 2, where the predicted noise from model $f_{\theta}$ is predicted after the update of the guiding channel, it is predicted before any update happens in Algorithm 3 

\item \textbf{Updating the Guiding Channel (Line 6):} The guiding channel update in Algorithm 3 involves the output of the model $f_{\theta}$, which is the predicted noise. This will change the noise schedule to follow a new predicted noise.

\begin{equation}
c'_{t-1} = \sqrt{1 - \beta_{t-2}} \cdot c'_{t-2} + \sqrt{\beta_{t-2}} \cdot f_{\theta}(z_t, t)
\end{equation}
\end{enumerate}

Compared with Algorithm 2, Algorithm 3 involves dynamically updating the noise schedule of the guiding channels. In practice, the guiding channels can be precalculated for all timestamps if using Algorithm 2, but in Algorithm 3, this process has to be repeated at each timestamp.

\section{Training Settings}

We use a U-Net model as the base diffusion model, consisting of approximately 168.9 million parameters, using the following settings:

\begin{itemize}
\item \textbf{Image size:} Images are 4x64x64 pixels.
\item \textbf{Training Batch Size:} 32.
\item \textbf{Epochs:} 300. An epoch is one complete pass through the entire training dataset.
\item \textbf{Gradient Accumulation Steps:} 1. This defines the number of steps for which to update the gradients at the same time.
\item \textbf{Optimizer:} AdamW
\item \textbf{Learning Rate:} 2e-5.
\item \textbf{Learning Rate Warmup Steps:} 500. This increases the learning rate linearly for the first 500 steps.
\item \textbf{Precision:} We use mixed precision `fp16' to speed up computations.
\end{itemize}

The models consist of convolutional blocks, time-embedding blocks, transformer blocks, and residual blocks, which provide pathways for the model to learn inter-channel relationships. 
\end{document}